\documentclass[letterpaper, 10pt, twocolumn]{article}
\usepackage{clean}
\shorttitle{Contact state observer}

\usepackage{cite}

\usepackage{graphicx}
\usepackage{subfig}
\usepackage{amsmath,amssymb,amsfonts}
\usepackage{algorithm2e}
\usepackage{xcolor}
\usepackage[colorlinks=true, linkcolor=black, urlcolor=cyan, filecolor=black, citecolor=black]{hyperref}
\usepackage{url}
\usepackage{amsmath,bm}
\usepackage{multicol, multirow, makecell} 
\usepackage{comment}
\usepackage{algorithm2e}

\title{Differentiable Compliant Contact Primitives \\[0.15cm] for Estimation and Model Predictive Control}
\author{Kevin Haninger$^1$, Kangwagye Samuel$^{2}$, Filippo Rozzi$^3$, Sehoon Oh$^2$, and Loris Roveda$^4$
\thanks{*This work was supported by the National Research Foundation of Korea (NRF) grant funded by the Korea government (MSIT) (No. 5120201213805) and the European Union's Horizon 2020 research and innovation programme under grant agreement No. 101058521 — CONVERGING.}
\thanks{$^{1}$Department of Automation, Fraunhofer IPK, Berlin, Germany
        {\tt\small kevin.haninger@ipk.fraunhofer.de}}%
\thanks{$^{2}$Department of Robotics and Mechatronics Engineering, DGIST, Daegu, 42988, Korea
        {\tt\small [ksamuel27, sehoon]@dgist.ac.kr}}
\thanks{$^{3}$Politecnico di Milano, Department of Mechanical Engineering, Milano, Italy
        {\tt\small filippo.rozzi@mail.polimi.it}}
\thanks{$^{4}$Istituto Dalle Molle di Studi sull'Intelligenza Artificiale (IDSIA), Scuola Universitaria Professionale della Svizzera Italiana (SUPSI), Università della Svizzera Italiana (USI) IDSIA-SUPSI, Lugano, Switzerland {\tt\small loris.roveda@idsia.ch}}%
}
\begin{document}

\maketitle

\begin{abstract}
Control techniques like MPC can realize contact-rich manipulation which exploits dynamic information, maintaining friction limits and safety constraints. However, contact geometry and dynamics are required to be known. This information is often extracted from CAD, limiting scalability and the ability to handle tasks with varying geometry. To reduce the need for a priori models, we propose a framework for estimating contact models online based on torque and position measurements. To do this, compliant contact models are used, connected in parallel to model multi-point contact and constraints such as a hinge. They are parameterized to be differentiable with respect to all of their parameters (rest position, stiffness, contact location), allowing the coupled robot/environment dynamics to be linearized or efficiently used in gradient-based optimization. These models are then applied for: offline gradient-based parameter fitting, online estimation via an extended Kalman filter, and online gradient-based MPC. The proposed approach is validated on two robots, showing the efficacy of sensorless contact estimation and the effects of online estimation on MPC performance.

\end{abstract}

\section{Introduction}
Many contact-rich tasks -- opening a door, screwing a lid onto a bottle, inserting a plug -- require robustness over variation in task parameters such as contact normal, object inertia, or contact stiffness. Robot control can be improved when these parameters are known or estimated, \textit{e.g.}, estimating a door's inertia can improve manipulation \cite{minniti2021}, and estimating environment stiffness can improve impedance control \cite{roveda2021}. Additionally, most contact-aware planning methods assume knowledge of contact geometry \cite{tedrake2020, carius2018}, which is often extracted from CAD \cite{huang2021}. 

\begin{figure}
    \centering
    \includegraphics[width=0.9\columnwidth]{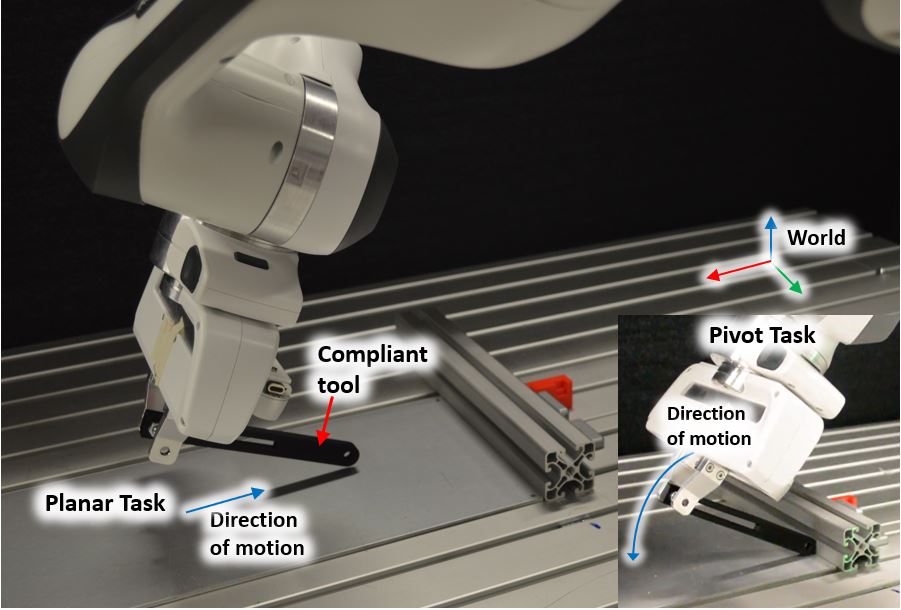}
    \caption{Robot interacting with the environment via a compliant tool, sliding along a surface of varying height and pivoting about a corner.}
    \label{fig:exp_franka_setup}
\end{figure}

The contact geometry can be directly applied as a kinematic constraint \cite{stewart2000}, \textit{i.e.}, assuming it is perfectly stiff. These models are common in contact planning \cite{mastalli2020, carpentier2021}, scaling to multi-point contact \cite{aydinoglu2022}. Alternatively, a compliant contact model can be used, either as a relaxation of the stiff dynamics \cite{pang2022}, or when the robot, tool, or environment has meaningful compliance \cite{vandermerwe2022}, for example as seen in Fig. \ref{fig:exp_franka_setup}. Compliant contact models have been effective for force control \cite{rossi2016, roveda2021}, model predictive control \cite{minniti2021, castro2022}, control of contact impacts \cite{wang2022}, and task monitoring \cite{kato2019, haninger2018}. 
The effective environment stiffness is application-specific, often requiring data-based identification. This can be done online. When force and position are measured, adaptive control  \cite{diolaiti2005, parigipolverini2020} or recursive least squares \cite{rossi2016, minniti2021}  can be used.  When force/torque (F/T) measurements from a sensor are not available, the environment stiffness can be estimated from the robot motor position and motor torque \cite{roveda2021, roveda2021a}, improving estimates of external force over a momentum observer \cite{deluca2005, garofalo2019}.

Another contact parameter is the contact geometry, such as contact Jacobian or signed distance function \cite{tedrake2020}. It is often assumed that the contact geometry is known, especially in locomotion \cite{carius2018}, or where it can be extracted from CAD data \cite{mastalli2020, huang2021}. However, in manipulation it may be desired to model contact without needing CAD, to enable interaction with natural world objects or simplify deployment. Contact geometry can also be estimated from data when F/T measurements are available \cite{popov2017} based on analytical models, which can also be done in collision detection to localize the collision \cite{haddadin2017a}.

These model-based techniques have good performance, but needing an a priori model limits the ease of applying them to new problems. Deep learning can be applied for contact-rich tasks \cite{elguea-aguinaco2023}, but direct deep learning of contact has been shown to have poor data-efficiency and generalizeability \cite{parmar2021}. A possible compromise is an expressive parameterized model which can be integrated to optimization and learning methods. Automatic differentiation (AD) is a standard tool for optimization, allowing the efficient computation of the derivative of complex functions composed of differentiable operations \cite{baydin2018}, applied in robotics to learn dynamics \cite{smith2018}, estimate friction parameters \cite{lelidec2021}, and in contact model predictive control \cite{mastalli2020}. 

\begin{figure}
    \centering
    \includegraphics[width=0.9\columnwidth]{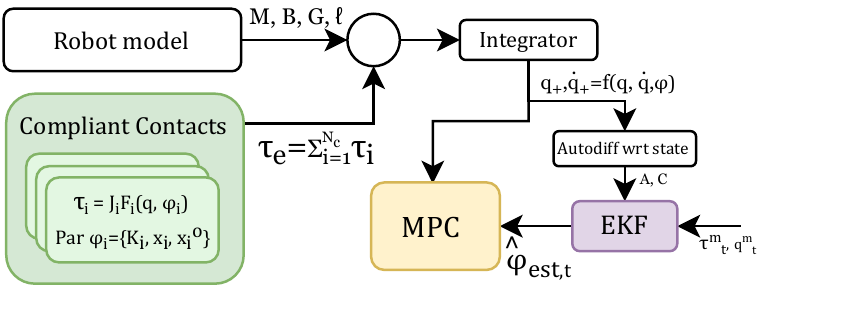}
    \caption{The proposed framework, where multiple compliant contact models are connected in parallel to the robot with parameters which can be fit offline or estimated online. These coupled dynamics are used to build an EKF and MPC, where the estimated parameters from the EKF update the model in the MPC online.}
    \label{fig:process}
\end{figure}

To improve the robustness of model-based control in contact, this paper proposes a framework for estimating contact models online based on torque and position measurements, where a compliant contact model which is differentiable with respect to all parameters (stiffness, normal, rest position, and contact location on robot) is developed. The coupled robot/environment dynamics are discretized, providing a differentiable dynamics which is here applied to: offline parameter fitting by gradient-based optimization, online parameter estimation with an extended Kalman Filter (EKF), and gradient-based MPC as seen in Fig.~\ref{fig:process}. This process is simplified by AD, which supports gradient-based optimization for parameter fitting, generation of Jacobian matrices, and linearizing the dynamic/observation equations.

This paper contributes a novel estimator for parameterized multi-point contact. The estimation of stiffness includes the spatial direction, extending 1-DOF stiffness estimators with a fixed direction \cite{roveda2021}. The paper also contributes an MPC with parametric model adaptation in contact, extending recursive least squares estimates in 1-DOF \cite{minniti2021}. This paper is structured as follows: first, the continuous-time robot and environment models are introduced in Section \ref{sec:models}, then the discrete system equations are derived in Section \ref{sec:disc_sys}. Offline parameter fitting and online estimation are presented in Section \ref{sec:par_fit}. Experiments with a collaborative robot in Section \ref{sec:exp} compare force and stiffness estimates with the momentum observer, and showing the feasibility of estimating contact geometry. All these estimation problems are derived from the same primitive, showing the flexibility of the approach.  

\subsection{Notation}
A series $\bullet_{1:t} = [\bullet_1,\dots,\bullet_t]$, vertical concatenation $[a;b]$ and horizontal $[a,b]$, the normal distribution is $\mathcal{N}(\mu,\Sigma)$, the likelihood of a normal variable is $\mathcal{N}(x|\mu, \Sigma)$, and the next time step is $\bullet_+ = \bullet_{t+1}$. The Jacobian of $f$  with respect to $x$ is denoted $D_x f$.

\section{Robot and Compliant Contact Model \label{sec:models}}
This section introduces the models for robot dynamics and contact models.
\subsection{Robot Dynamics}
We assume the standard serial manipulator robot dynamics 
\begin{equation}
    M(q)\ddot{q}+ C(q,\dot{q}) + B\dot{q} + G(q) = \tau_m + J^T(q)F_e \label{inertial_dyn}
\end{equation}
and forward kinematics of
\begin{align}
x = \ell(q), \quad \dot{x} = J(q)\dot{q}, \quad \ddot{x} = J(q)\ddot{q} + \dot{J}(q)\dot{q} \label{fwd_kin}
\end{align}
are available in an AD framework, with joint position $q\in\mathbb{R}^n$, inertia matrix $M(q)$, Coriolis terms $C(q, \dot{q})$, viscous damping $B$, gravitational torque $G(q)$, joint motor torque $\tau_m$, force at the tool-center point (TCP) $F_e$, TCP pose $x$, and standard TCP Jacobian matrix $J$. We consider the pose $x=[p, R]$ with position $p\in\mathbb{R}^3$ and rotation matrix $\mathbb{R}^{3\times 3}$, denoting the Jacobian $D_qp = J_p$. We simplify notation with the torque error as
\begin{eqnarray}
    \tilde{\tau} = \tau_m - C(q, \dot{q}) - G(q).
\end{eqnarray}

\subsection{Compliant Contact Model}
\begin{figure}
    \centering
    \includegraphics[width=0.7\columnwidth]{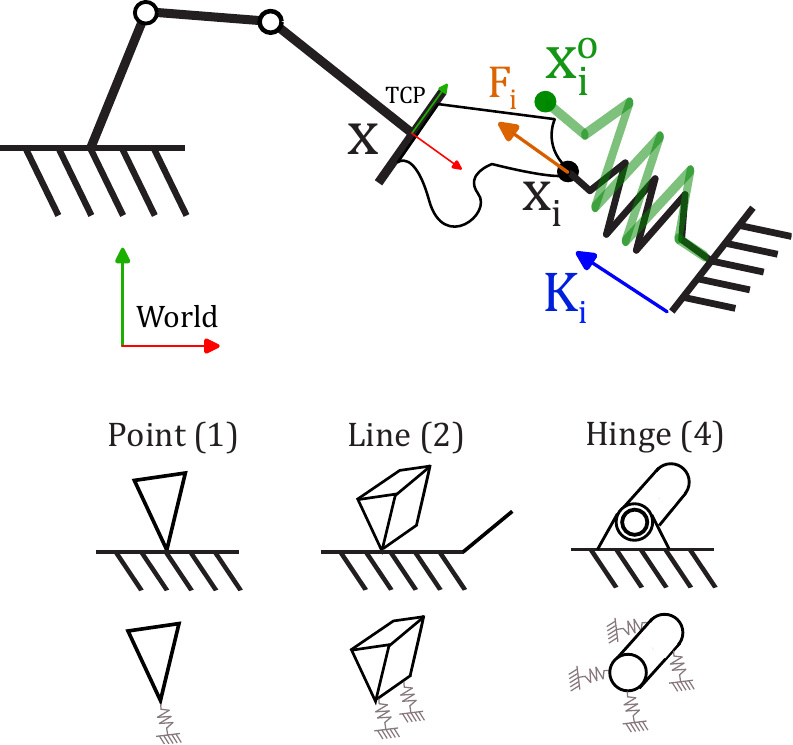}
    \caption{Stiffness contact model, where a contact point exerts only a normal force at a point on the robot. The geometry between the contact point and the end effector is assumed to be unknown but fixed. To model more complex kinematic constraints, multiple contact stiffnesses can be added in parallel. }
    \label{fig:contact_model}
\end{figure}
Consider a contact model as seen in Fig.~\ref{fig:contact_model}, where the contact force $F_i$ at the point of contact is expressed as 
\begin{equation}
    F_i = \mathrm{diag}(K_i)(x_i^o-x^w_i),  \label{env_contact}
\end{equation}
where stiffness $K_i\in\mathbb{R}^3$ has a rest pose at $x_i^o\in\mathbb{R}^3$, and the contact point $x^w_i(q)\in\mathbb{R}^3$ position in world coordinates is given by a fixed position in TCP frame $x_i$, i.e. $x^w_i(q) = p(q)+R(q)x_i$. The contact normal and stiffness are jointly described by the vector $K_i\in \mathbb{R}^3$, \textit{i.e.}, the contact normal is $n_i = K_i/\Vert K_i \Vert$, and the stiffness is $\Vert K_i\Vert$. 

Denoting the contact Jacobian $J_i = D_q(x_i^w) = D_q(p + Rx_i)$, the joint torques can be written as
\begin{eqnarray}
\tau_i &=& J_i^TF_i \\
 & = &(J_p^T + D^T_qRx_i)\mathrm{diag}(K_i)(x^o_i-p-Rx_i).
\end{eqnarray}
The contact Jacobian $J_i \in\mathbb{R}^{3\times n}$ maps the joint speeds to velocity at contact point $x_i^w$, where the position Jacobian of the TCP $J_p$ is recovered if $x_i=0$, i.e. the contact point is at the TCP.
 
This model is parameterized by $\left[ K_i, x_i, x^o_i \right]$, where each is unconstrained in $\mathbb{R}^3$. When the contact normal $n_i$  and a scalar stiffness is used in the parameterization \cite{castro2022}, the norm $\Vert n_i \Vert = 1$ and stiffness $>0$ must be maintained.  

We denote the parameters of the contact model which will be differentiated as $\phi$. The vector $\phi$ includes the values to be estimated online or fit offline, and the dynamics can be built to include the parameters relevant for a specific application. 

\subsection{Multi-point Contact Model}
Applying multiple single DOF contacts, as seen in the bottom of Fig.~\ref{fig:contact_model}, can be used to represent higher-order constraints. When $N_c$ stiffnesses are connected in parallel, their contribution to joint torque adds, resulting in total contact torque $\tau_e$ of  
\begin{eqnarray}
        \tau_e = \sum_{i=1}^{N_c} J_i^TF_i = J_p^T\sum_{i=1}^{N_c}F_i + \sum_{i=1}^{N_c}D^T_q(Rx_i)F_i\label{total_contact}.
\end{eqnarray}

\section{Discretized Linearized Dynamics  \label{sec:disc_sys}}
This section derives the discretized dynamics and linearizes them.
\subsection{Integrator \label{sec:integrator}}
As typical in contact models, we use semi-implicit integration which handles stiff differential equations better \cite{stewart2000, castro2022}. For a time step of $h$, denoting the next time step of a variable $\bullet_+$ and dropping the argument $q$,
\begin{eqnarray}
    q_+ & = & q + h\dot{q}_+ \label{integ_pos} \\
    \dot{q}_+ & = & \dot{q} + hM^{\texttt{-}1}\left(\tilde{\tau} + \tau_e - B\dot{q}_+\right) \nonumber \\
    & = & (I+hM^{-1}B)^{-1}\left(\dot{q} + hM^{\texttt{-}1}\left(\tilde{\tau} + \tau_e \right)\right).\label{integ_vel}
\end{eqnarray}

We note that $(I+hM^{-1}B)^{-1}=I-h(M+hB)^{-1}B$ \cite[(157)]{pedersen2016}, and simplify the dynamics as
\begin{eqnarray}
    q_+ & = & q + h\dot{q} + h^2\delta \label{integ_pos2} \\
    \dot{q}_+ & = & \dot{q} + h\delta \\
\delta & = & \texttt{-}(M\texttt{+}hB)^{\texttt{-}1}B\dot{q} \texttt{+} (I\texttt{+}hM^{\texttt{-}1}B)^{\texttt{-}1} M^{\texttt{-}1}\left(\tilde{\tau} \texttt{+} \tau_e \right),\nonumber
\end{eqnarray}
where $\delta$ is the impulse caused by damping terms, torque error and contact torque. Note this impulse simplifies to $\delta = M^{-1}(\tilde{\tau}+\tau_e)$ if $B=0$. As $\tau_e$ depends on the contact primitive parameters $\phi$ and $q$, the dynamics can be written as 
\begin{equation}
    \begin{bmatrix} q_+ \\ \dot{q}_+\end{bmatrix} = f\left(\begin{bmatrix} q \\ \dot{q}\end{bmatrix}, \tilde{\tau}, \phi \right). \label{eq:nonlin_dyn}
\end{equation}

\subsection{Linearized Dynamics}
For an EKF, the dynamics \eqref{eq:nonlin_dyn} must be linearized with respect to the state. When $N_e$ contact parameters are to be estimated online, these parameters are denoted $\phi_{est} \in \mathbb{R}^{3 N_e}$, resulting in a total state vector
\begin{equation}
    \xi= \begin{bmatrix} q; \dot{q}; \phi_{est} \end{bmatrix}.
\end{equation}

The system dynamics of \eqref{eq:nonlin_dyn} can then be linearized to find
\begin{eqnarray}
    \xi_+ & \approx & \begin{bmatrix} I + h^2D_q\delta & hI & h^2 D_{\phi_{est}}\delta\\ h D_q\delta &  \left(I+hM^{-1}B\right)^{-1} & hD{\phi_{est}} \delta\\
    0 & 0 & I \end{bmatrix}\xi+ b \nonumber\\
                                                   & \approx & A \xi + b + w \label{eq:lin_dyn},
\end{eqnarray}
with process noise $w\sim\mathcal{N}(0, Q)$, bias terms $b$, $D_q\delta\in\mathbb{R}^{n \times n}$ and $D_\phi\delta \in \mathbb{R}^{n\times 3N_e}$. In \eqref{eq:lin_dyn}, the dependence of $\tau_e$ on $q, \dot{q}$ is ignored to simplify the notation - these terms are automatically calculated in the AD framework.

\section{State and Parameter Estimation with Primitives \label{sec:par_fit}}
The parameters $\phi$ depend on the environment and task and may need to be fit offline or estimated online. Parameters that are fit offline, as introduced in Section \ref{sec:offline_fit}, are denoted $\phi_{fit}$. The fit parameters update the primitives and are set to their optimized numerical values at runtime. Then, any parameters to be estimated online $\phi_{est}$ are estimated with the EKF as introduced in Sec. \ref{sec:online_estimator}.

\subsection{Offline Parameter Fitting \label{sec:offline_fit}}
An advantage of the AD framework is that gradient-based optimization can be more easily implemented without needing to put the problem in a standard form such as least-squares. When fitting contact model parameters based on only robot measurements, the full state $\xi$  is unobserved. We apply a simplified expectation maximization approach \cite{dempster1977, durbin2012} to iteratively estimate i) the full state trajectory $\xi_t$ with the Kalman filter from Section \ref{sec:online_estimator} to produce state mean and covariance $[\mu_{1:T}, \Sigma_{1:T}]$, then ii) the model parameters are optimized  assuming $\xi_{1:T} = \mu_{1:T}$. 

In the model parameter optimization step, the dynamic model is fit to minimize the least-squares prediction error as
\begin{align}
    \phi_{fit} = \arg \min_{\phi} &  \sum_{t=1}^T\Vert \mu_{t+1} - f(\mu_t,\tau_{m,t}, \phi) \Vert  \label{eq:fit} \\ & + 0.5\Vert x^w_{i,t}-x_i^o\Vert+ \beta_{K_i}\Vert K_i\Vert_1 + \beta_{x_i} \Vert x_i \Vert \nonumber,
\end{align}
where $\beta_\bullet$ is regularization ($\beta_{K_i} = 1e-9$, and $\beta_{x_i} = 5$), and $x^w_{i,t}$ is the contact position $x^w_i$ in world coordinates at time $t$. 

\subsection{Online EKF \label{sec:online_estimator}}
The state and parameters $\phi_{est}$ can be jointly estimated in an EKF. The EKF uses observations of joint positions, and optionally joint torques, to estimate the complete state. For the state at time $t$, $\xi_t^T=[q^T_t, \dot{q}^T_t, \phi^T_{est, t}]$, and observations $y^m_t$, denote the posterior belief as
\begin{equation}
    p(\xi_t | y^m_{1:t}) = \mathcal{N}(\mu_t, \Sigma_t). \label{belief}
\end{equation}
The observations of joint position and torque, $q^m$ and $\tau^m$ are given by 
\begin{align}
    q^m & = C_q\xi + v_q \label{eq:obs_model} \\
    \tau^m & =  \tau_m + v_m \\
    & \approx C_\tau \xi + v_m,
\end{align}
with measurement matrices $C_q=[I_n, 0_n, 0_{3N_e}]$, $C_\tau = [D_q\tau_e, 0_n, 0_{3N_e}]$, process noise $v_q\sim\mathcal{N}(0,R_q)$  and $v_\tau \sim \mathcal{N}(0, R_\tau)$  assumed to be independent and identically distributed, $I_n$ is an identity matrix of dimension $n$, and $0_n$ a zero matrix with $n$ columns.

When only the joint positions are measured $y^m = q^m$, the EKF uses observation matrix $C=C_q$ and $R=R_q$, and joint torque $\tau_m$ is treated as an input, i.e. motor torque.  When both the joint positions and torques are measured, $y^m = [q^m; \tau^m]$, the matrices become $C=[C_q; C_\tau]$  and $R=\mathrm{diag}(R_q, R_\tau)$.   The process noise $Q$ is composed of $Q = \mathrm{diag}(Q_q, Q_{\dot{q}}, Q_\phi)$,  where depending on which parameter is being estimated, a different $Q_\phi$ is used.

An EKF is then implemented in joint space with belief space update of \cite{thrun2002}
\begin{eqnarray}
    \bar{\Sigma} & = & A\Sigma A^T + Q \nonumber \\ 
    L & = & \bar{\Sigma} C^T \left( C \bar{\Sigma} C^T + R\right)^{-1} \\
    \mu_+ & = & Ly^{m}_+ + (I - LC)f(\mu, \tau^m) \label{mean_update} \\
    \Sigma_+ & = & (I - LC)\bar{\Sigma}, \label{cov_update}
\end{eqnarray}
where $\Sigma$ is the error covariance matrix of the estimate, $L$ is the Kalman gain and $A$ is from \eqref{eq:lin_dyn}.

\subsection{Observability of Online Contact Estimation}
When estimating contact parameters of \eqref{env_contact} online, the question arises if $\phi_{est}$  can be reliably estimated.  One way to verify this is the observability of the state $\xi$.  While the system is nonlinear, we can verify local observability via linearized matrices $A$ in \eqref{eq:lin_dyn} and $C_q$ in \eqref{eq:obs_model}, simplifying with damping $B=0$ and dropping the $h^2$ terms in $A$ as
\begin{align}
    \mathcal{O}  =\begin{bmatrix} C_q \\ C_qA \\ \vdots \\ C_qA^{n-1}\end{bmatrix} 
     = \begin{bmatrix} I & 0 & 0 \\ I & hI & 0 \\ I+h^2D_q\delta & 2hI & 3h^2 D_{\phi}\delta \\
     I+3h^2D_q\delta & 3hI + h^3 D_q\delta & 5h^2 D_{\phi}\delta \end{bmatrix},
\end{align}
 To check the rank of the observability matrix $\mathcal{O}$, we first do row elimination, yielding
\begin{align}
    \mathcal{O} = & \begin{bmatrix} 
    I & 0 & 0 \\ 
    0 & hI & 0 \\ 
    -5h^2D_q\delta & 3h^3D_q\delta & 0 \\ 
    h^2D_q\delta & 0 & 3h^2 D_{\phi}\delta \end{bmatrix}.
\end{align}
From this we see that a sufficient condition is that $D_\phi\delta$  is of full column rank, where if $M(q)$ is not singular and $B=0$, this is the condition
\begin{equation}
    \mathrm{rank}(D_\phi\tau) = 3N_e
\end{equation}  
For example, this Jacobian can be written for $D_{x_i^o}\tau_e=J_i^T\mathrm{diag}(K_i)$. From this, we can see that estimating two rest positions $x_1^o$ and $x_2^o$  requires that $J_1^T\mathrm{diag}(K_1)\neq J_1^T\mathrm{diag}(K_1)$, where if $x_1=x_2=0$ we have the condition that $K_1 \neq K_2$; that the contacts must be in different directions. The observability conditions can be checked for a specific estimation problem with the help of the AD framework. 
 
 \section{MPC with Compliant Contact}
As the contact model is differentiable, it can also be used for gradient-based control methods such as model predictive control (MPC). Thus, the same model can be used for estimation and control, where the parameters can be updated online from the estimator. 

\subsection{Impedance Dynamics}
We assume a robot with a Cartesian impedance controller is used, providing desired joint torques of
\begin{eqnarray}
 \tau_{m} & = -J^T_p\left(K_{imp}(p-x^{d})+D_{imp}J_p\dot{q}\right)
\end{eqnarray}
where $x^d\in\mathbb{R}^3$ is the virtual rest position of the impedance spring, $K_{imp}\in\mathbb{R}^{3\times 3}$ the virtual stiffness and $D_{imp} \in\mathbb{R}^{3\times 3}$ the damping. Both $K_{imp}$ and $D_{imp}$ are diagonal and are tuned for the application, the only control variable is $x^d$.

\subsection{Multiple Shooting Problem}
A multiple-shooting MPC problem is written with the dynamics of \eqref{eq:nonlin_dyn},
\begin{eqnarray}
 x^d_{t:t+H} & = & \arg \min_{x^d_{t:t+H}} \sum_{i=t}^{t+H}c(q_i, \dot{q}_i, x^d_i, \hat{\phi}_{est,t}) \\
  & \mathrm{s.t.}\,\, &\begin{bmatrix} q_{i+1} \\ \dot{q}_{i+1} \end{bmatrix} = f(q_i, \dot{q}_i, \hat{\phi}_{est,t}) \\
& & g(q_i, \dot{q}_i, x^d_i) > 0
\end{eqnarray}
where $\hat{\phi}_{est, t}$ is the estimate of parameters $\phi_{est}$ at time step $t$. We use a cost function 
\begin{eqnarray}
 c(q, \dot{q}, x^d, \phi) = \Vert p - p^d \Vert + Q_v \Vert J \dot{q} \Vert + \sum_i Q_f \Vert F^d - F_i \Vert
\end{eqnarray}
where $Q_v$, $Q_f$ are weights to adjust the velocity and force tracking terms, and $F^d$ is a desired contact force. While achieving a desired force $F^d$ can be more simply achieved with a force controller, this setup allows force tracking with a higher impedance, which can be advantageous in tasks where higher impedance robot behavior is needed (e.g. positioning). The constraint $g$ limits the effective force of the impedance controller as 
\begin{eqnarray}
    g(q, x^d) = \overline{F}_{imp}-\Vert K_{imp}(p - x^d) \Vert_2 
\end{eqnarray}
where  $\overline{F}_{imp}$ is a force limit for the impedance controller.  

\section{Experimental Validation \label{sec:exp}}
This section describes the implementation and experimental validation of the proposed approach.  First, the problem is applied to sensorless estimation of environment parameters, where only joint positions are measured.  Then, estimation and control is applied on a robot where both joint position and torque are measured.  
The software is built on the AD framework CasADi \cite{andersson2019}. For the fitting of parameters and the MPC, IPOPT \cite{wachter2006} is used.  The robot dynamics and kinematics are built in Pinocchio \cite{carpentier2019} with CasADi support. The inertial model available from the manufacturers is used and modified to include the motor inertia on the diagonal elements. The code and experiment data are available at \url{https://gitlab.cc-asp.fraunhofer.de/hanikevi/contact_mpc}.

\begin{figure}
    \centering
    \includegraphics[width=0.475\columnwidth]{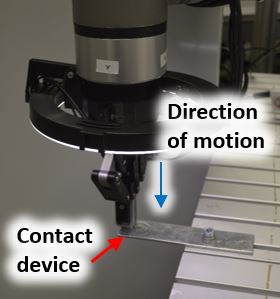}
    \caption{The vertical contact experiment for sensorless estimation.\label{fig:exp_setup}}
\end{figure}

\subsection{Sensorless Estimation \label{sec:sensorless_eval}}
A Universal Robots UR16e robot is brought into vertical contact with the environment as seen in Fig.~\ref{fig:exp_setup}.  The flange F/T sensor is used for validation of estimated forces, and motor position and current are measured at $500$ Hz. The gearbox ratios $[101, 101, 101, 54, 54, 54]$ and the motor torque constants $[0.119, 0.119, 0.098, 0.107, 0.107, 0.107]$ are used to translate the motor current measurements into $\tau_m$.  The parameters used are $Q_q = 1e\texttt{-}1 I$, $Q_{\dot{q}} = 1e4I$,   $R_q = 5e\texttt{-}2I$, and viscous damping of $B=0.2I$.

To see the observer accuracy, three observers are compared: the EKF with offline fit $K_i$,  the EKF with $\phi_{est} = K_i$, and the momentum observer. The robot makes vertical contact as seen in Fig.~\ref{fig:exp_setup}  with  $105$ N of contact force in the vertical direction by jogging the robot. The data is recorded and played back offline to give the same reference input to each observer.  

\subsubsection{Momentum Observer and Stiffness Estimator}
To benchmark the force and stiffness estimation, a momentum observer \cite{deluca2005, garofalo2019} is used to estimate the TCP forces, finding the residual as $ r_{t+1} = K_O\left(M_t\dot{q}_t-h(r_t - \tau_{e,t})\right)$,
where $h$ is the time step, $M_t, \tau_{e,t}$ are from Section \ref{sec:integrator}, and $K_O=20$ the observer gain. The residual was translated to TCP forces by $F_t^{mo} = (J^T)^+r_t$, where $\bullet^+$ is the pseudoinverse.  The estimated force $F_t^{mo}$ is used with the TCP position $x_t$ to estimate the external stiffness $K_t^{mo}$ as 
\begin{eqnarray}
 K_t^{mo} = \frac{F_{t+W}^{mo}-F_{t}^{mo}}{\min\left(\vert x_{t+W}-x_{t}\vert, 5e-4\right)},
\end{eqnarray}
where $W$ is a smoothing window length, and the denominator is modified to prevent division by zero.

\subsubsection{Force Estimate}
\begin{figure}
    \centering
    \subfloat[Estimated force in X, Y, and Z directions]{\includegraphics[width=0.85\columnwidth]{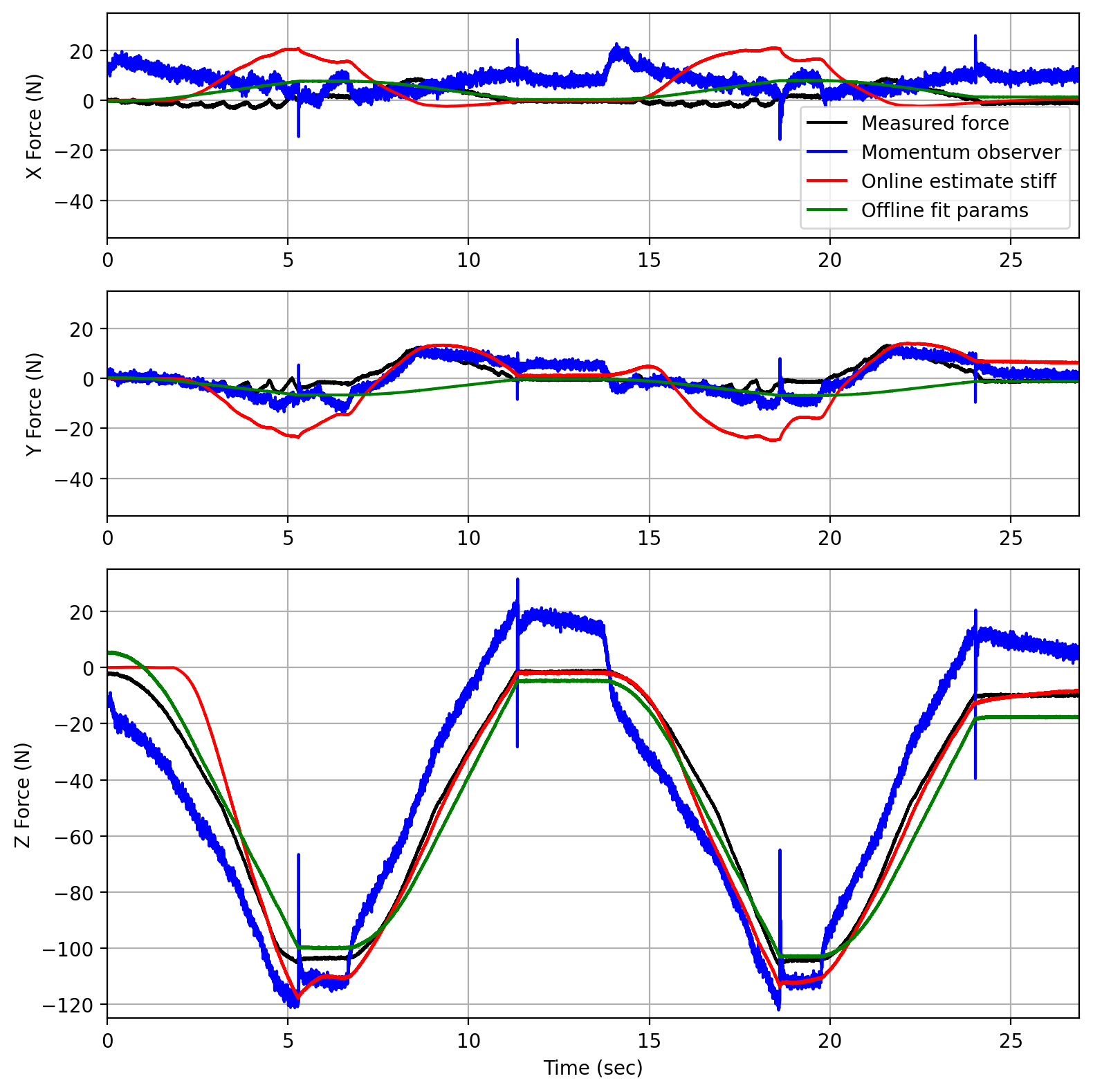}} \\
    \caption{Comparison of estimated and measured force in the Z-contact task, forces shown in TCP coordinate frame}
    \label{fig:force_comparison}
\end{figure}

The three observers are compared in their ability to estimate the TCP force, shown in Fig.~\ref{fig:force_comparison}, compared with the directly measured force. It can be seen that the momentum observer (blue) has high-frequency electrical noise, low-frequency error in the gravitational model, and discontinuities when the motion is stopped. Its major advantage is simplicity and independence from environment dynamics \cite{garofalo2019}.

On the other hand, the EKF observers have reduced high-frequency noise and low-frequency error. An improvement is to be expected as they are including more information - and are therefore environment-specific. The EKF with offline stiffness estimate (green) presents a smaller low-frequency error in the Z direction, and any discontinuities from the motor command are removed as this force is estimated over the robot state (joint position). The online estimate (red) also has a lower error in the Z direction, but a substantial lag in the estimate in the force (around $2.6$ seconds from contact).  This could not be addressed by tuning noise parameters or initial covariance, it is suspected that it may be influenced by off-diagonal elements in the covariance which are not initialized. The online EKF stiffness estimate performs comparably to the momentum observer in the X/Y direction, where there is minimal motion. 

\subsubsection{Stiffness Estimate \label{sec:stiff_est}}
\begin{figure}
    \centering
    \subfloat[Estimated stiffness in the Z base coordinate]{\includegraphics[width=0.85\columnwidth]{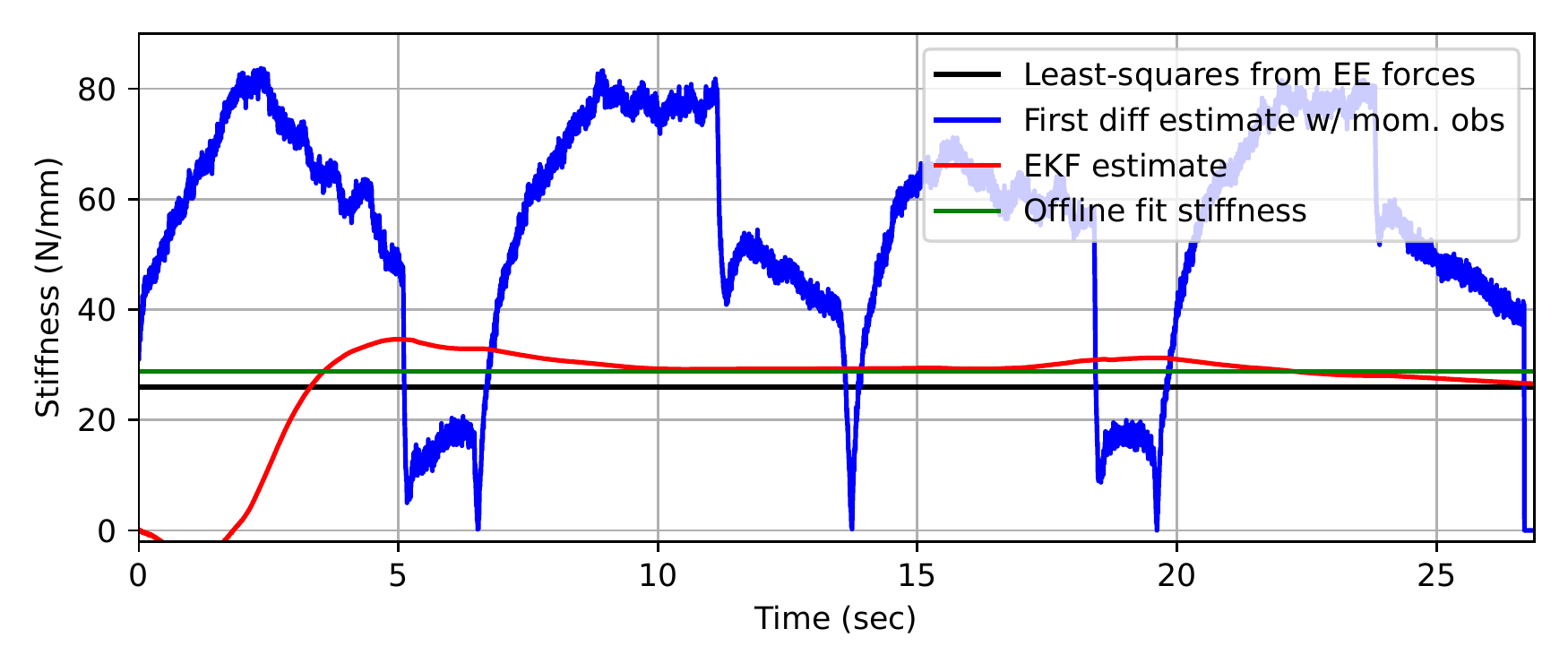}} \\
    \subfloat[Estimated stiffness in X/Y coordinates]{\includegraphics[width=0.85\columnwidth]{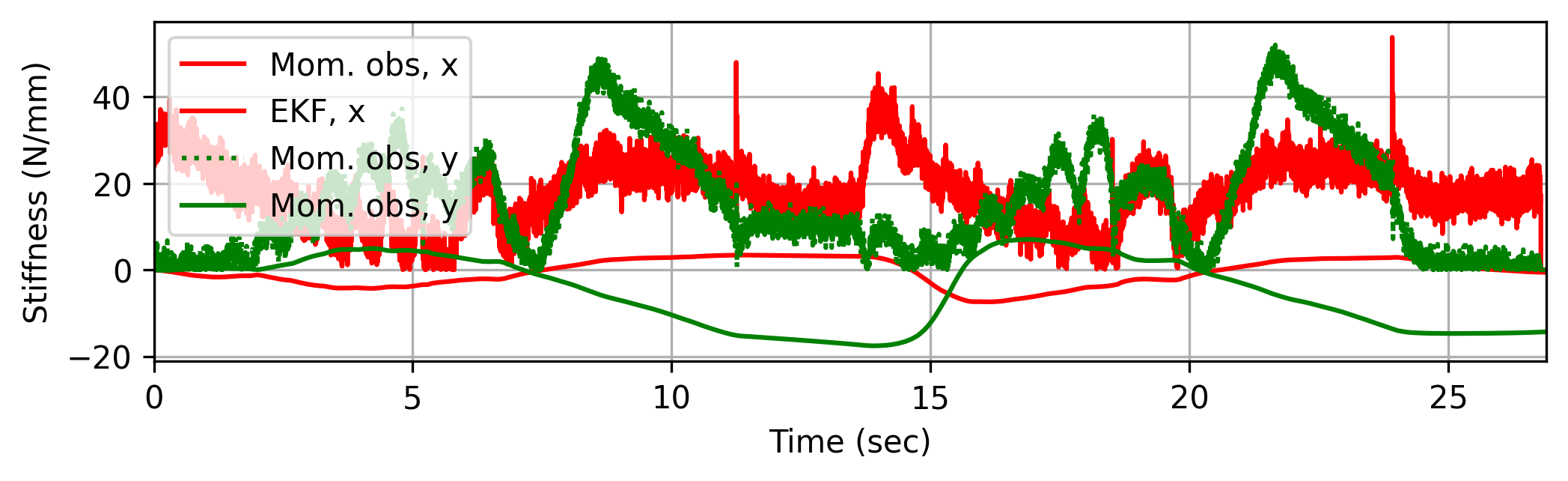}} \\
    \caption{Vertical contact experiments with estimated stiffness in Z (a) and X/Y (b)}
    \label{fig:stiff_comparison}
\end{figure}

The observers are then compared in their ability to estimate the stiffness online, shown in Fig.~\ref{fig:stiff_comparison}. Here, the offline estimates provided by the method in Sec. \ref{sec:offline_fit} (green, $28.3$ N/mm) are also compared with the least-squares fit of stiffness from the F/T sensor measurements and TCP pose (black, $25.8$ N/mm). 

It can be seen in Fig.~\ref{fig:stiff_comparison}(a) that the offline estimate of the environment stiffness is a reasonable approximation of the directly estimated stiffness, without needing the use of an F/T sensor.  On the other hand, the online estimate of stiffness has more variation, taking approximately $2.6$ seconds to rise to converge, and the magnitude varies as the direction of motion varies. The momentum observer estimate of stiffness is found with a moving-average window of $W=50$ time steps or $0.01$ seconds but remains noisy. It also has a much higher degree of variation during the task, also having large jumps when the residual of the momentum observer jumps due to the discontinuity in motor current. 

In the X and Y directions, the EKF estimates a much lower stiffness, whereas the momentum observer-based approach has higher average estimates of stiffness.  

\subsection{Sensored Estimation}
When the joint torque measurements $\tau^m$ are available, the estimates can be substantially improved.  We test this on the hardware shown in Figure \ref{fig:exp_franka_setup}. We test the ability to estimate $K_i$, $x_i^o$ and $x_i$  online for a single contact point. The $Q$ parameters from \ref{sec:sensorless_eval} are used with $Q_\tau=5I$, and $R_{K_i} = I$, $R_{x_i} = R_{x_i^o} = 1e-6$.  These experiments can be seen in the attached video.  

When jointly estimating $\phi_{est} = [x_i^o, x_i]$ , the force prediction error remains low, but the estimates diverge from the TCP.  Estimates of the stiffness can distinguish between different contact materials and detect a change in the normal direction. 

\subsection{Online MPC}
To investigate the value of online estimation of contact parameters, we use two experiments, the first moving along a plane of uncertain vertical location ($N_c = 1$), then pivoting about a hinge location ($N_c=2$). In both cases, the performance is compared from using offline estimates of the stiffness models and online estimates. 

\subsubsection{Planar task}
In this task, the robot should move along the plane keeping $3$ N contact force in $Z$.  An MPC is used with and without online estimation of the rest position $x_i^o$ with a single contact model where $K_1=[0,0, 2570]$. The MPC problem is set up with $H=13$, $h=0.03$, $p^d = [0.35, -0.35, 0.01]$, $Q_f=5e-5$,  $Q_v = 0.05$, $F^d=3$, and $\overline{F}_{imp} = 15$.  

\begin{figure}
    \centering
    \includegraphics[width=\columnwidth]{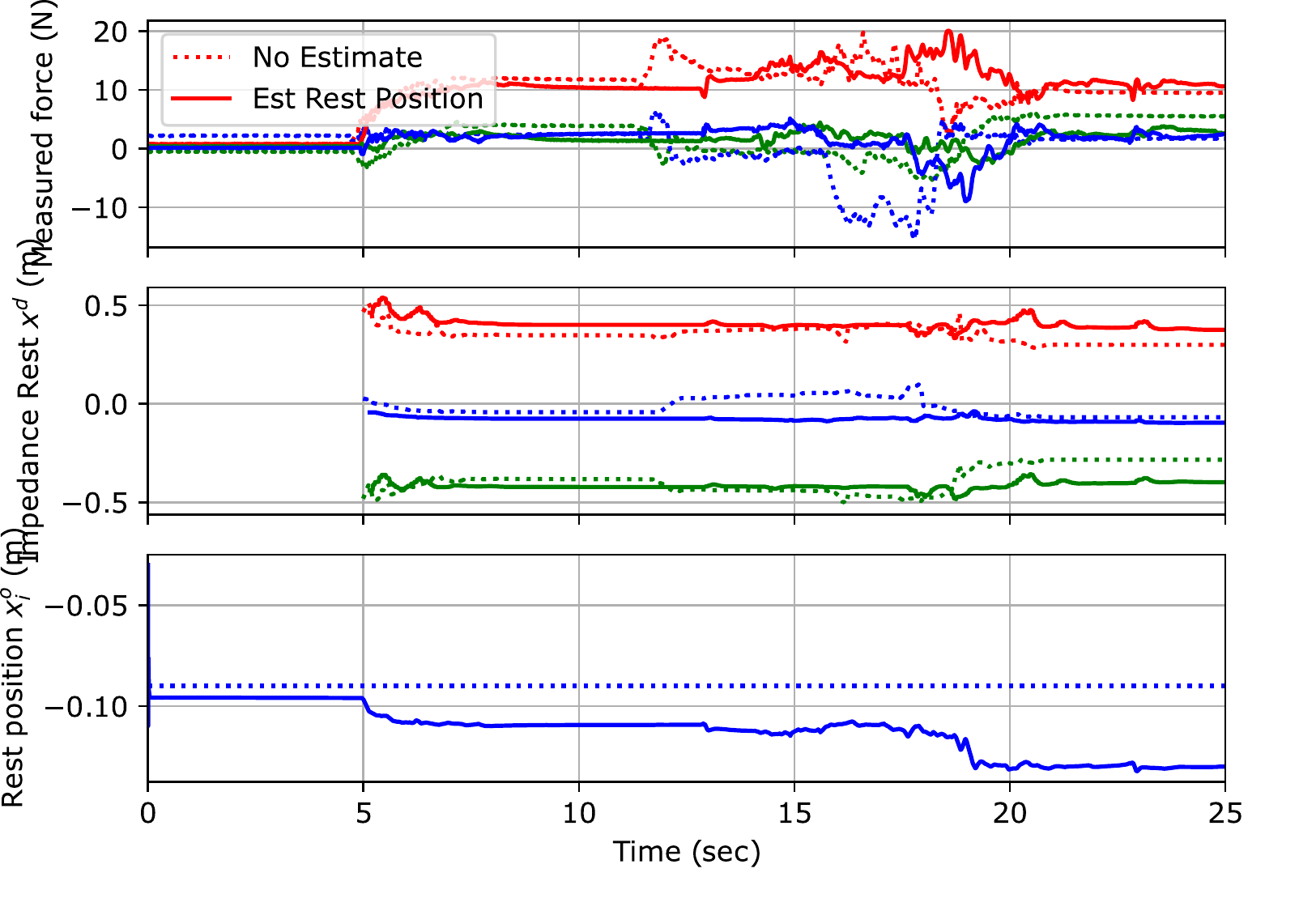}
    \caption{Plane problem, showing the measured forces, impedance position, and estimated rest position with and without online EKF estimation, where red, green and blue indicate $x, y, z$ as shown in Figure \ref{fig:exp_franka_setup}}
    \label{fig:exp_plane}
\end{figure}

The results can be seen in Fig.~\ref{fig:exp_plane}, where the MPC is started at 5 seconds.  It can be seen that the MPC without the estimation has a higher variation in $Z$ forces, whereas with online estimation variation in plane height is estimated and compensated. 

\subsubsection{Pivoting task}
In this task, the MPC problem is set up with $H=13$, $h=0.03$, $p^d = [0.32, -0.6, -0.15]$, $Q_f=2e-5$,  $Q_v = 0.05$, $F^d=15$, and $\overline{F}_{imp} = 30$. Two contacts are applied, with a stiffness of $K_1 = [0, 0, 2570]$ and $K_2 = [3300, 0, 0]$. 

\begin{figure}
    \centering
    \includegraphics[width=\columnwidth]{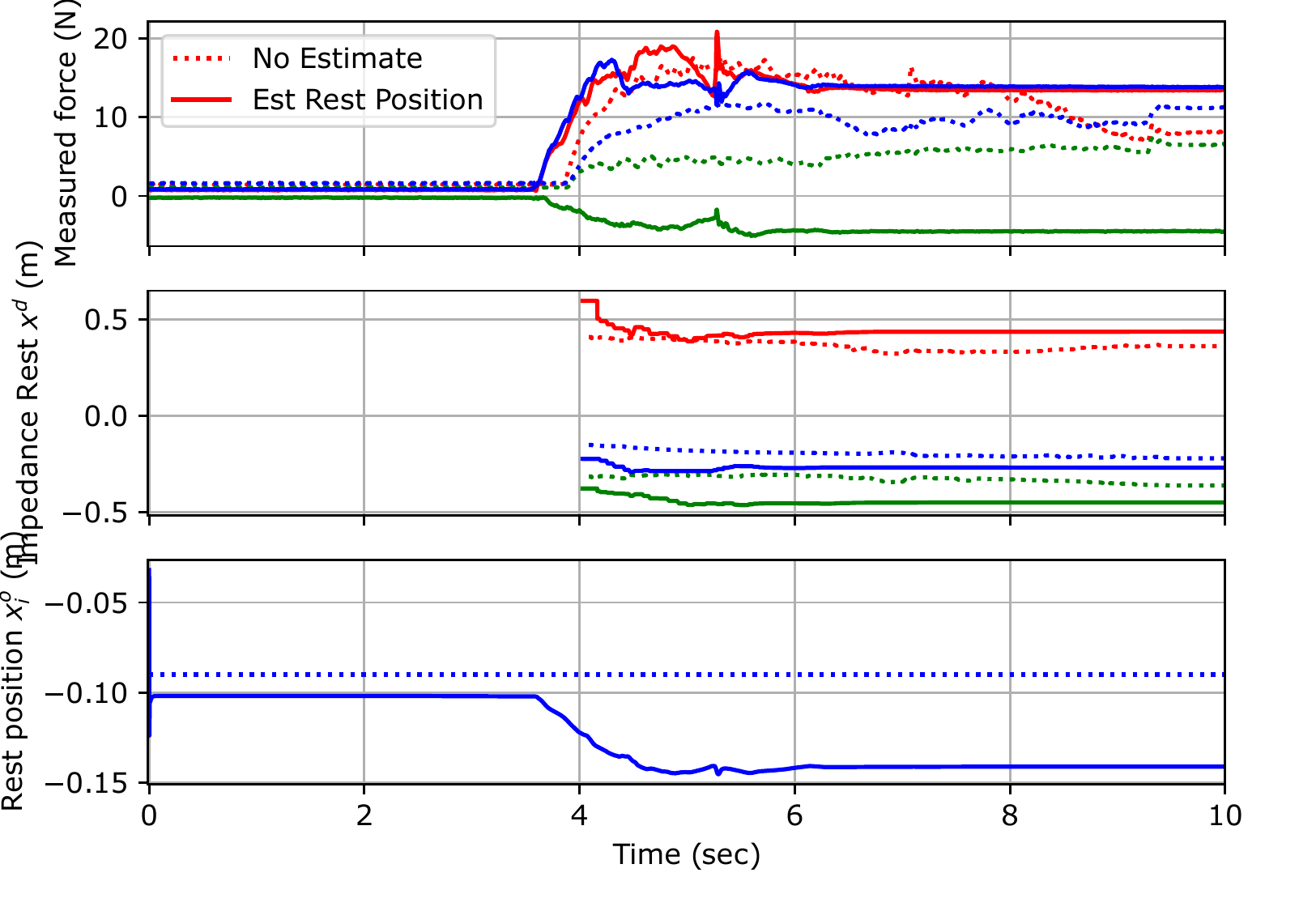}
    \caption{Pivot problem, showing the measured forces, impedance position, and estimated rest position with and without online EKF estimation, where red, green and blue indicate $x, y, z$ as shown in Figure \ref{fig:exp_franka_setup}}
    \label{fig:exp_pivot}
\end{figure}

The result with and without online estimation of the contact rest positions $x_1^o$ and $x_2^o$ can be seen in Fig.~\ref{fig:exp_pivot}. It can be seen that the online estimation results in good steady-state tracking of the desired force in $Z$ and $X$, whereas the few centimeters of error result in poorer tracking for the no estimation case.

\section{Conclusion}
This paper demonstrated that differentiable compliant contact parameters can support offline fitting, online estimation and control in a unified framework. This methodology was verified in sensorless estimation of external force and stiffness, as well as estimation problems when joint torque are measured. The approach was then shown to improve MPC performance, allowing online adaptation to parametric uncertainty in rest position for contact tasks. 


\ifdefined\ARXIV
    \printbibliography
\else
    \bibliographystyle{IEEEtran}
    \bibliography{lib_diff}
\fi

\end{document}